\documentclass{article}

\usepackage{microtype}
\usepackage{graphicx}
\usepackage{subfigure}
\usepackage{booktabs} 

\usepackage{hyperref}

\usepackage[accepted]{icml2025}

\usepackage{amsmath}
\usepackage{amssymb}
\usepackage{mathtools}
\usepackage{amsthm}

\usepackage[capitalize,noabbrev]{cleveref}

\usepackage{tabularx}
\usepackage{graphicx} 
\usepackage{float}
\usepackage{placeins}
\theoremstyle{plain}

\theoremstyle{definition}

\theoremstyle{remark}

\usepackage[textsize=tiny]{todonotes}

\icmltitlerunning{EMO2: End-Effector Guided Audio-Driven Avatar Video Generation}

\begin{document}

\twocolumn[
\icmltitle{EMO2: End-Effector Guided Audio-Driven Avatar Video Generation}




\begin{icmlauthorlist}
\icmlauthor{Linrui Tian}{}
\icmlauthor{Siqi Hu}{}
\icmlauthor{Qi Wang}{}
\icmlauthor{Bang Zhang}{}
\icmlauthor{Liefeng Bo}{}

{Institute for Intelligent Computing, Alibaba Group}\\
{\{tianlinrui.tlr, husiqi.hsq, wilson.wq, zhangbang.zb, liefeng.bo\}@alibaba-inc.com}
\url{https://humanaigc.github.io/emote-portrait-alive-2/}

\end{icmlauthorlist}






\vskip 0.3in
]




\begin{abstract}
In this paper, we propose a novel audio-driven talking head method capable of simultaneously generating highly expressive facial expressions and hand gestures. Unlike existing methods that focus on generating full-body or half-body poses, we investigate the challenges of co-speech gesture generation and identify the weak correspondence between audio features and full-body gestures as a key limitation. To address this, we redefine the task as a two-stage process. In the first stage, we generate hand poses directly from audio input, leveraging the strong correlation between audio signals and hand movements. 
In the second stage, we employ a diffusion model to synthesize video frames, incorporating the hand poses generated in the first stage to produce realistic facial expressions and body movements. Our experimental results demonstrate that the proposed method outperforms state-of-the-art approaches, such as CyberHost~\cite{cyberhost} and Vlogger~\cite{vlogger}, in terms of both visual quality and synchronization accuracy. This work provides a new perspective on audio-driven gesture generation and a robust framework for creating expressive and natural talking head animations.
%

\end{abstract}
\section{Introduction}

\begin{figure}[h]
  \centering
  \includegraphics[width=0.5\textwidth]{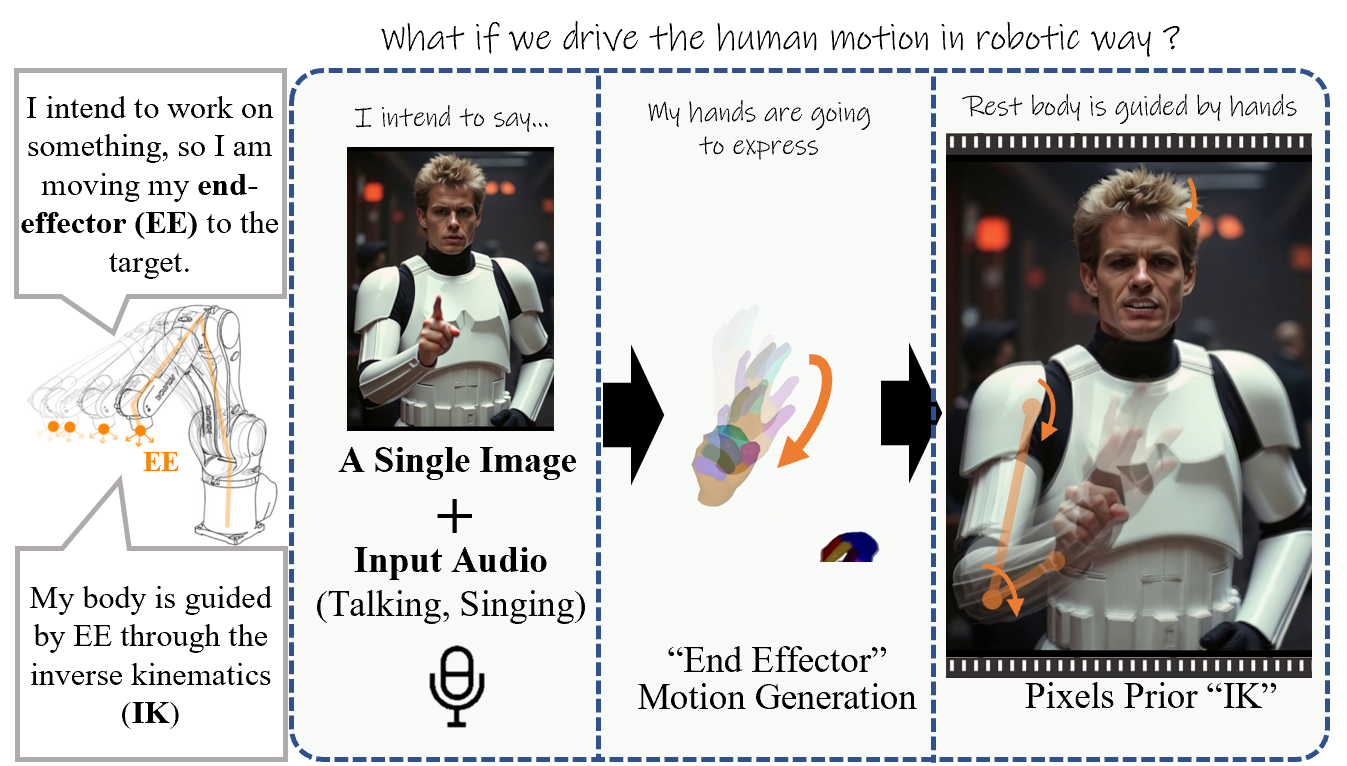}
  \caption{The motivation behind our method. Human motion, similar to that of robots, involves planning the "end-effector" (EE), typically the hands, towards the target position. The rest of the body then cooperates accordingly with the EE, abiding by inverse kinematics principles.
  }
  \label{fig:motivate}
\end{figure}

Audio-driven human video generation, which aims to create synchronized facial expressions and body gestures, remains a critical research area with a wide range of applications. While there have been noteworthy achievements in generating synchronized facial expressions from audio~\cite{tian2025emo} and human-centric videos~\cite{hunyuanvideo}, challenges persist in creating audio-synchronized human videos that exhibit richly vivid motions, particularly in the realm of co-speech video generation. Some methods~\cite{vlogger, cyberhost, he2024cospeechgesturevideogeneration} have attempted to address co-speech video generation, but they often fall short in reproducing rich body motion or lack the generalization ability necessary for diverse scenarios.

One of the fundamental challenges in this domain stems from the complexity of the human body, which operates as a sophisticated multi-joint articulated system~\cite{humanbody}. This complexity makes it particularly difficult to predict rich gestures that are synchronized with audio, especially within pixel space. Several approaches~\cite{SMPL, vlogger, talkshow, cocogesture} have sought to define human body structure by focusing on predicting full-body or half-body poses, using either joint angles~\cite{diffsheg, talkshow} or positions~\cite{hit-dvae, idrees2024advmt}. However, these approaches struggle with capturing the intricate coupling of multiple joints in natural human motion, particularly in speech-related activities, leading to suboptimal results. A key observation in our work is that the correlation between audio and different body joints varies significantly.

We observed that robotic control systems such as manipulators and humanoid robots are often designed to mimic human behavior. As illustrated in  \autoref{fig:motivate}, robotic systems typically define tools or grippers at the end as the \textbf{"end-effector"} (EE). Tasks for robots are often defined as steering the EE's 6-DoF (degree of freedom) posture because the EE's motion provides a more straightforward and intuitive task representation. This simplification eases the control process compared to managing each robot joint angle separately. In some robotic implementations, including manipulation~\cite{tian2021vote} and motion planning~\cite{nmpc}, robot joint poses are often determined through Inverse Kinematics (IK).

This principle inspires us to redefine the target for co-speech human motion generation. Hands, which act as the "EE" in daily life, are more directly tied to human intention and exhibit a stronger relationship with audio than other body joints. For instance, when speaking or singing, individuals frequently produce intentional hand gestures that align with their expressive content.

However, obtaining joint angles through IK can sometimes result in singularities, leading to suboptimal motion planning solution. To address this issue, some robotic methods~\cite{rakita2018relaxedik} incorporate prior knowledge into IK to ensure plausible joint solutions. Similarly, recent 2D pixel generation models~\cite{ldm, animatediff, hunyuanvideo} have shown the integration of human body structure knowledge, suggesting that pixel generation models may implicitly incorporate insights about human IK. Thus, we could potentially use hand motion to represent partial body movement while leveraging 2D generative models' capability to generate other body parts, which we refer to as \textbf{"pixels prior IK"}. This approach enables the reproduction of entire characters, synchronization of audio with lip movements, and preservation of proper body structure, ultimately resulting in coherent and continuous co-speech videos.

Inspired by this insight, we propose a two-stage framework for audio-driven gesture and facial expression generation. In the first stage, we focus on mapping audio to hand poses, capitalizing on the strong correlation between audio and hand movements. This approach simplifies the problem by reducing the complexity of the mapping space and allows for more precise control over gesture generation. In the second stage, we employ a diffusion-based model to synthesize video frames, incorporating the generated hand poses to produce realistic facial expressions and body movements.

By leveraging these insights, our method not only generates coherent and continuous co-speech videos but also ensures proper synchronization of audio with lip movements and realistic body dynamics.

In summary, the main contributions of our work are as follows::
\begin{itemize}
    \item We identify the strong correlation between audio and hand movements and propose a simplified, two-stage framework for audio-driven gesture generation.
    \item We introduce a diffusion-based model for synthesizing realistic facial expressions and body movements from generated hand poses.
    \item We demonstrate through experiments that our method outperforms state-of-the-art approaches in terms of visual quality, synchronization accuracy, and motion diversity.
\end{itemize}




\section{Related Work}
\label{sec:related}
\textbf{Co-speech gesture generation.}  

Over the years, co-speech gesture synthesis, like expression synthesis, has evolved from rule-based methods to data-driven approaches. Rule-based methods implement a hard mapping from speech to gestures by linking phonemes to specific gesture patterns \cite{rule-based-co-speech-hand-generation1, rule-based-co-speech-hand-generation2, rule-based-co-speech-hand-generation3}. Data-driven methods explore various neural network architectures to learn the connection between speech and gestures from extensive training data. For instance, Faceformer \cite{faceformer2022} employs a transformer to learn audio-to-expression connections, Codetalker \cite{codetalker} incorporates a learned codebook to strengthen connections with discrete motion priors, Talkshow \cite{talkshow} introduces a speech-gesture paired dataset and implements an autoencoder for facial motions along with a compositional VQ-VAE for body and hand motions, while \cite{co-speech-ges-gen-gan} utilizes a GAN-based pipeline.

Recently, inspired by the remarkable success of diffusion-based methods in image and video generation, motion generation approaches have begun to incorporate diffusion processes. DiffGesture \cite{diffgesture} and DiffSHEG \cite{diffsheg} developed diffusion processes for skeleton and SMPL \cite{SMPL} pose sequences, using a transformer to attend to audio or other modality inputs. Cocogesture \cite{cocogesture} trained an audio control-net to integrate audio embeddings into gesture features. It is noteworthy that most methods initially generate parameters from predefined body structures such as SMPL and then convert them into gesture movements in 3D coordinates, with the final gesture qualities being constrained by the simplified body structure. Few methods attempt to implement a diffusion process directly on hand postures.

\textbf{Audio-driven human video generation.}

Recent works~\cite{hunyuanvideo, HaCohen2024LTXVideo, yang2024cogvideox} have significantly advanced human-centric video generation. However, research on audio-driven human video generation often focuses primarily on talking heads. EMO~\cite{tian2025emo} introduces an audio-driven video diffusion framework for creating expressive talking head videos. Loopy~\cite{loopy} incorporates longer sequences of historical information to enhance video vividness. Hallo~\cite{hallo, cui2024hallo2, cui2024hallo3} explores methods to enhance audio control. Some studies aim to animate both the head and body motion with audio. Vlogger~\cite{vlogger} develops a two-stage method that uses audio to drive a 3D human representation, followed by animation from this representation, but it shows insufficient correlation with audio input. Cyberhost~\cite{cyberhost} directly drives body movement with audio in pixel space. Echomimic~\cite{echomimicv2} animates partial body parts using pre-extracted control signals while driving other parts with audio. He et al.~\cite{he2024cospeechgesturevideogeneration} utilize latent motion signals for body movement.

\section{Method}
\label{sec:method}

\subsection{Overview}
Given a single reference image of a character, our approach can animate the character by inputting a musical/vocal audio clip, preserving the natural facial expression and body motion in harmony with the variation of the fed audio. 

\begin{figure*}[tb]
  \centering
  \includegraphics[width=\textwidth]{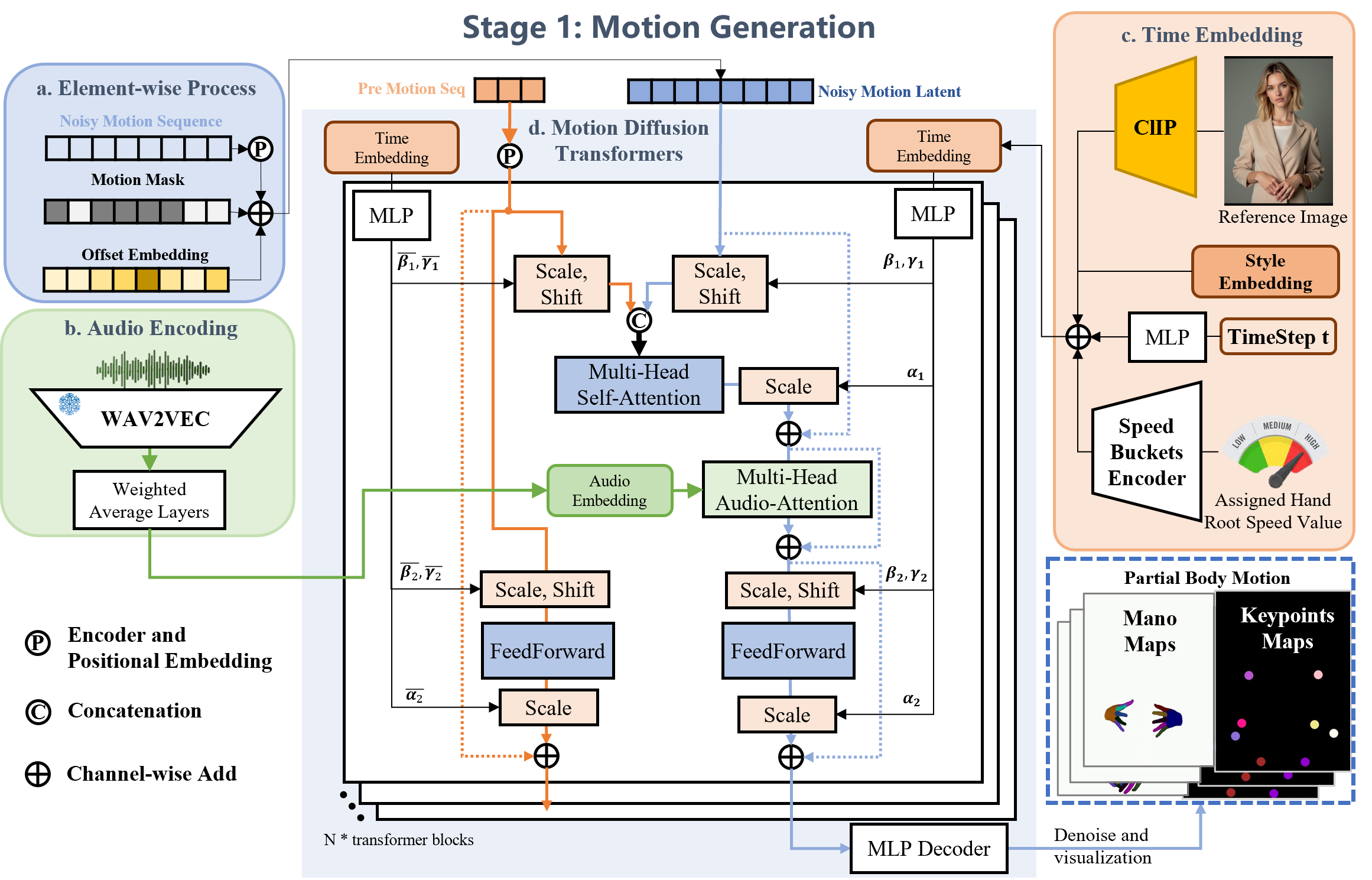}
  \caption{Overview of the stage 1 hand motion generation framework. The framework includes serveral  DiT blocks as backbone. Audio embeddings are injected via cross-attention, style and speed embeddings are added on timestep, previous motion latent sequence is concatenated on current noisy motion latent sequence for smooth transition. Hand masks that mask out invisible hands frames are directly added on noisy motion latent.
  }
  \label{fig:pipeline1}
\end{figure*}

We propose a two-stage method to drive body animation from a single image using audio. In the first stage, as illustrated in  \autoref{fig:pipeline1}, we develop a motion diffusion model where audio inputs are translated to synchronized partial body motion signals. These generated motion signals are then utilized in the second stage, as shown in  \autoref{fig:pipeline2}. The second stage employs a ReferenceNet-based diffusion architecture \cite{animate_anyone, tian2025emo}, guided by both the audio and motion signals, animating the reference image into co-speech video.

\subsection{Preliminaries}
Our two-stage methodology is based on the framework of diffusion models (DDPM) \cite{ddpm}, which assume a forward noising process where noise $\epsilon \sim N(0,1)$ is gradually applied to real data $x_0$, resulting in noisy data $x_t$ at a specific timestep $t$. The diffusion model is trained to learn the reverse process, aiming to remove the noise $\epsilon$ from the noisy data $x_t$. During this denoising process, control signals $c$, such as audio and mapping data in our method, are introduced to achieve the desired outcome $x_0$. The training objective for this denoising process is defined as:
\begin{equation}
\mathcal{L}=\mathbb{E}_{t,c, x_t, \epsilon}\left[ ||\epsilon - \epsilon_{\theta}(x_t, t, c)||^2\right]
\end{equation}
where $\epsilon_{\theta}$ denotes the diffusion models. In Stage 1, our method implements $\epsilon_{\theta}$ using a transformer-based structure \cite{transformers}, discussed in Section \ref{sec::partial_body}. Stage 2 employs the ReferenceNet \cite{googletryon} backbone, which is an evolution of the Latent Diffusion Model (LDM) \cite{ldm}. It utilizes a Variational Autoencoder (VAE) \cite{vae} to map the original image feature distribution $x_0$ into a latent space $z_0$ to reduce computational load, while using ReferenceNet to blend reference image features with the latent features. In Stage 2, $\epsilon_{\theta}$ represents the Denoising U-Net,  discussed in Section \ref{sec::video_gen}. 


  \subsection{Partial Body Motion Generation}
\label{sec::partial_body}
Given audio inputs, instead of directly operating on pixels like EMO\cite{tian2025emo} does with facial expressions, a partial body motion generation model is devised to produce co-speech gestures as the intermediate driving signal for generating the human body in videos. The rationale is that human body movements are naturally articulated and can be represented by predefined gestures using 3D models such as SMPL for the body and MANO\cite{MANO} for the hands. These models provide unified parameters and facilitate easy control over body motion, but they compromise on expressiveness and diversity. Interestingly, we discover that using hand movements alone suffices to describe upper body motion, as the movements of other body parts can be derived through Inverse Kinematics (IK). Unlike the IK used in robotics—which often encounters issues with singular solutions, necessitating robots to be hardcoded or trained with prior knowledge to avoid singularities—IK in human video generation is effectively managed. This is because the video generation backbone is pretrained with a large amount of human body structure data in the form of pixels, which provides robust prior knowledge for resolving human IK challenges. Therefore, in the motion generation stage, we only generate co-speech hand motions as the driving signal for the next stage of video generation. Experiments show that the video generation model can produce reasonable upper body motions given only the control signal of the hands. Additionally, releasing control over the arms and other parts of the body further increases the expressiveness and intensity of upper body motion, compared to existing motion generation methods such as DiffSHEG, Talkshow and CoCoGesture that operate on all joints of the upper body.

To generate movement of the hands, we adopt the diffusion transformer (DiT)\cite{DiT} as the backbone due to its scalability to large-scale datasets. The model consists of multiple DiT blocks, each performing self-attention on noisy motion latents and cross-attention between audio features and noisy motion latents. Inspired by Pixart-\(\alpha\)\cite{pixartalpha}, \textit{AdaLN-single} is implemented to inject timestep embeddings into each DiT block. With the attention mechanism, coherent MANO hand coefficients consistent with speech rhythms are generated. To ensure smooth transitions between clips, the last frames of the motion sequence in the previous clip are concatenated to the current motion sequence. The audio features are extracted through Wav2vec\cite{wav2vec}. Aside from hand movements, predefined keypoints of the human upper body are included as weak supervision for video generation.

\textbf{Hand Motion Mask and Hand Offset.} In some frames of training dataset, the MANO hand annotations are inaccurate or missing because the hands are covered or out of sight in the frame. To filter out the influence of these bad cases, motion masks indicating the frames of valid  annotated MANO hand parameters are used and added to the noisy motion latents. Motion masks also indicate the padding length of the sequence. An offset embedding that describes the relative position and rotation of the body is also added to disentangle hand motion from root body posture differences in the dataset.

\begin{figure*}[tbh]
  \centering
  \includegraphics[width=0.8\textwidth]{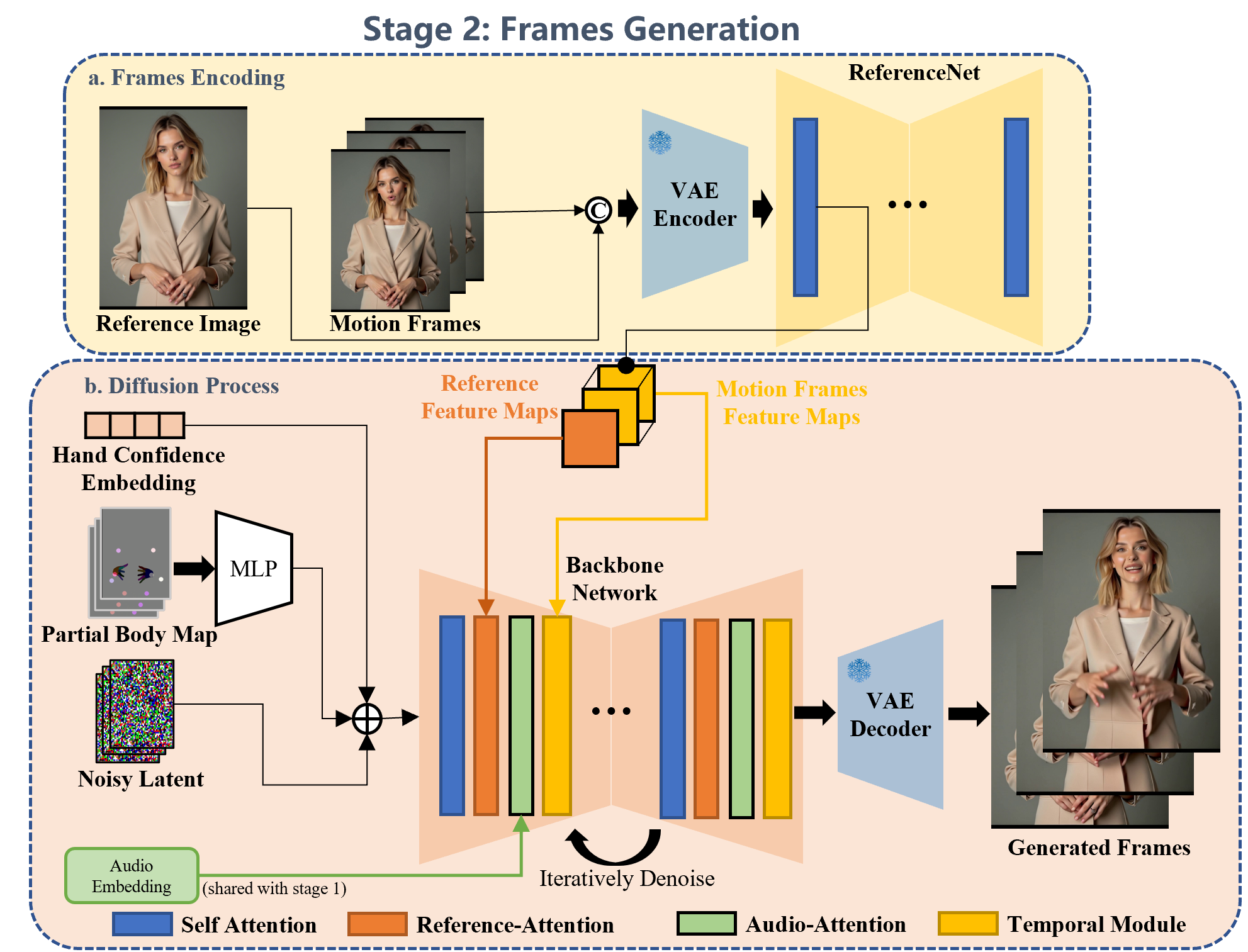}
  \caption{
The overview of the Stage 2 video generation pipeline, which is based on the Parallel Reference Network structure. The ReferenceNet extracts visual features from both the reference image and motion frames. The MANO maps and keypoint maps generated in Stage 1 are passed through the denoising Backbone Network to guide the character's motion. Additionally, trainable hand confidence embeddings enhance the quality of the generated hands. The audio embeddings are injected to ensure synchronization between audio and visual elements.
  }
  \label{fig:pipeline2}
\end{figure*}

\textbf{Style, Speed, and Reference Image Embedding.} For generating hand motion in different styles like singing, speaking, and gesture dance, a style embedding can be added to the timestep embedding. In addition, we filtered the hand movements into different speed buckets, similar to EMO, where each bucket has a center and a radius, and a specific speed value is encoded based on its distance to each bucket. Speed embeddings are then added to the timestep to control the speed of each hand respectively. In practice, we found that using the variance of the hand translation to represent movement amplitude, rather than speed, can achieve more pronounced control over hand movement. Optionally, the information of a reference image can be injected into the backbone to generate motion more suitable for the reference image context, such as when there are objects like a guitar or microphone in hand. Given a reference image, we adopt the vision model from CLIP\cite{clip} to obtain the encoded class embedding of the image and add it to the timestep, thus enabling the model to generate hand motion accordingly.

\subsection{Co-Speech Video Generation}
\label{sec::video_gen}

The design of our video generation framework is based on the EMO\cite{tian2025emo}. As shown in Figure \ref{fig:pipeline2}, Our Backbone Network is fed with multi-frame noise latent input, and tries to denoise them to the consecutive video frames during each time step. The framework can be divided into four parts: 1) \textbf{Denoising}: The Backbone Network is a denoising 2D-UNet integrated with temporal modules from AnimateDiff\cite{animatediff}; 2) \textbf{Frames Reference}: To maintain the character ID, we deploy the ReferenceNet parallel to the Backbone, it inputs the reference image and motion frames\cite{tian2025emo} to get the 2D image features. Those features are injected into the Backbone through cross attention in the spatial and temporal dimensions respectively; 3) \textbf{Audio-Driven}: To drive the character by audio, the audio features shared with the stage 1 are integrated through cross-attention; 4) \textbf{Motion Guidance}: The MANO maps and keypoint maps generated in Stage 1 are concatenated channel-wise and integrated with latent features to modulate body motion. 

\FloatBarrier
\textbf{Hand Motion Control.} 
The MANO maps produced in stage 1 guide the character's motion. They explicitly describe the hand movement in the generated frames, detailing aspects such as shape, size, and pose. However, even with these explicit control signals, poor MANO annotation in training dataset can lead to bad hand representations. Similar to the design of hand clarity control \cite{cyberhost}, we employ the hand confidence scores of MANO hand detection in training frames. These scores may decrease in situations of significant occlusion or motion blur, serving as conditional input to enhance the quality of generated hands. Specifically, we multiply a trainable embedding by these confidence scores to create a hand confidence embedding, which are directly added to the latent features. During inference, assigning higher confidence scores results in clearer and structurally correct hands.

\begin{table*}[htb]
  \centering
  \caption{The Quantitative comparisons with other motion generation methods.}
  \label{tab:quantitative_comp_motion}
  \setlength{\tabcolsep}{2pt}
  \begin{tabularx}{\textwidth}{l>{\centering\arraybackslash}X>{\centering\arraybackslash}X>{\centering\arraybackslash}X>{\centering\arraybackslash}X>{\centering\arraybackslash}X>{\centering\arraybackslash}X>{\centering\arraybackslash}X>{\centering\arraybackslash}X}
    \toprule
    Method & DIV$\uparrow$ & BA$\uparrow$ & PCK$\uparrow$ & FGD$\downarrow$ & DIV(smpl)$\uparrow$ & BA(smpl)$\uparrow$ & PCK(smpl)$\uparrow$ & FGD(smpl)$\downarrow$  \\
    \midrule
    Talkshow & 0.0961 & 0.6743 & 0.7872 & 0.0329 &  0.0400 &  0.6769 & 0.9529 & 0.4170  \\
    Diffsheg& 0.0158 & 0.7198 & 0.8300 & 0.03676 &  0.0306 &  0.7312 & 0.7635 & 4.2189  \\
    Ours on SMPL& 0.0886 & 0.7290 & 0.8163 & 0.0301 &  0.1442 &  0.7285 & 0.7345 & 4.5746  \\
    Ours on MANO& 0.1345 & 0.7626 & 0.8126 & 0.0373 &  - &  - & - & -  \\
    \bottomrule
  \end{tabularx}
\end{table*}

\begin{figure*}[htb]
  \centering
  \includegraphics[width=0.7\textwidth]{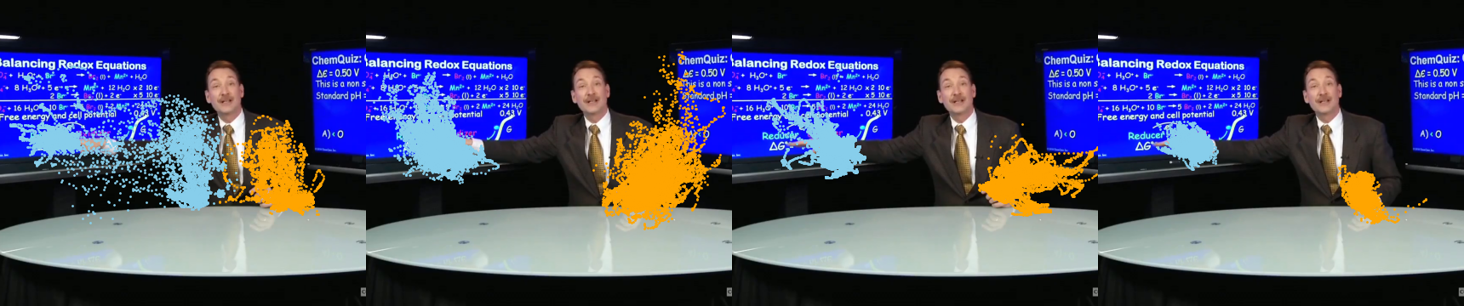}
  \caption{ The distribution of the generated hand positions from co-speech gesture generation methods based on Talkshow dataset. From left to right: Ours MANO based,Ours SMPL based, Talkshow, Diffsheg. 
  }
  \label{fig:hand distribution}
\end{figure*}

\textbf{Keypoints Guidance Control.} 
Similar to EchoMimicV2 \cite{echomimicv2}, our initial implementation utilized hand-only control signals, allowing other body parts to synchronize with audio signals and hand movements. However, we observed that large movements of the MANO hand signals were often incompatible with static torso, leading to unnatural performance in the videos. To address this issue, we introduce joint keypoints to supplement the motion-driven approach. The keypoints maps indicate the 2D positions of joints in the arms and legs. 
It is important to note that since the stage 1 model is trained using only 2D keypoint annotations, challenges arise in accurately maintaining 3D body structures, such as arm length. Consequently, the generated keypoints may not precisely define the entire body. However, they still serve as effective indicators of joint movement trends. Specifically, we apply large-kernel median filtering to the 2D keypoints along the temporal dimension during the training of the stage 2 model. This filtering approach forces the keypoints to intentionally misalign with the body joints, allowing the generated body movements to be guided by these points while still permitting the model some creative freedom. As a result, this strategy facilitates the generation of dynamic performance videos.

\textbf{Pose Discriminator.} 
Moreover, to enhance the character body structure, we implement a pose discriminator during the training stage. Specifically, at each training timestep, we utilize one-step sampling to compute the latent prediction \(z_{t0}\) from the model's output \(z_t\) at timestep \(t\). This latent result \(z_{t0}\) is then passed through the pose discriminator, which predicts the body pose keypoints and limb heatmap \(\hat{H}\). The pose discriminator loss \(L_{pd} = \| H - \hat{H} \|_2\) is incorporated into the denoising loss during training, where \(H\) represents the ground truth heatmap. The pose discriminator is based on ResNet \cite{resnet} and is pretrained in the latent space.
\FloatBarrier
\section{Experimets}
\label{sec:exp}

\subsection{Implement Details}

We train our two generation models individually. In the first part of hand motion generation, 24 DiT blocks with a hidden size of 512 are concatenated to form the backbone. The MANO hand model is used to describe hand movements, which includes 48 joint rotation values for each hand in axis-angle representation and 3 translation values. To eliminate the ambiguity in rotation, the 48 axis angles are converted to 64 quaternion parameters, yielding a total of 134 parameters for both hands. Hand parameter sequences are padded to 300 frames for arbitrary length generation, and the previous sequence of 12 frames is used for a smooth transition from the previous clip to the current one. We trained the first part of the model on an A100 GPU with a batch size of 8 for 400k training steps from scratch.

\begin{table*}[tb]
  \centering
  \caption{Quantitative comparisons with various body animation methods. The * symbol indicates evaluations performed on CyberHost's 12 demo videos, while ** denotes evaluations conducted on the Vlogger's 30 demo videos from its homepage.}
  \label{tab:quantitative_comp}
  \setlength{\tabcolsep}{2pt}
  \begin{tabularx}{\textwidth}{l>{\centering\arraybackslash}X>{\centering\arraybackslash}X>{\centering\arraybackslash}X>{\centering\arraybackslash}X>{\centering\arraybackslash}X>{\centering\arraybackslash}X>{\centering\arraybackslash}X>{\centering\arraybackslash}X>{\centering\arraybackslash}X}
    \toprule
    Method & FID$\downarrow$ & FVD$\downarrow$ & SSIM$\uparrow$ & PNSR$\uparrow$ & Sync-C$\uparrow$ & EFID$\downarrow$ & HKC$\uparrow$ & HKV$\uparrow$ & CSIM$\uparrow$ \\
    \midrule
    EchoMimicV2 & 33.42 & 217.71 & 0.662 & 65.13 &  4.44 &  1.052 & 0.425 & 0.150 & 0.519 \\
    MimicMotion& 25.38 & 248.95 & 0.585 & 64.09 &  2.68 &  0.617 & 0.356 & 0.169 & 0.608 \\
    w/o motion gen& 21.07 & 102.19 & 0.751 & 67.88 &  4.59 &  0.224 & 0.461 & 0.175 & 0.683 \\
    w/o hand confidence& 25.82 & 134.14 & 0.659 & 64.47 &  4.11 &  0.200 & 0.537 & 0.191 & 0.666 \\
    Ours & 27.28 & 129.41 & 0.662 & 64.62 &  4.58 &  0.218 & 0.553 & 0.198 & 0.650 \\
    \cmidrule(lr){1-10} 
    CyberHost* & - & - & - & - &  4.54 &  - & 0.723 & 0.107 & 0.708 \\
    Ours* & - & - & - & - &  4.70 &  - & 0.723 & 0.150 & 0.746 \\
    \cmidrule(lr){1-10} 
    Vlogger** & - & - & - & - &  1.94 &  - & 0.611 & 0.068 & 0.491 \\
    Ours** & - & - & - & - &  5.11 &  - & 0.604 & 0.154 & 0.565 \\
    \bottomrule
  \end{tabularx}
\end{table*}

In the second part of our approach, the video generation model training is divided into two stages. The first stage focuses on image training, where two frames are sampled and cropped to a resolution of \(704 \times 512\) pixels, serving as the reference frame and the target frame. During this stage, we optimize the ReferenceNet, the motion guidance layer, and the basic modules within the Backbone Network. The second stage involves audio-video training, during which the temporal modules, audio attention layers, and additional components are integrated into the model for optimization. For memory efficiency, the ReferenceNet is frozen in this stage. Each video clip comprises 24 frames at a resolution of \(704 \times 512\) pixels, with the number of motion frames set to 12. Both stages utilize 4 A100 GPUs, with batch sizes of 32 and 4 for the image and audio-video training stages, respectively. Each stage undergoes 100k training steps with a learning rate of \(1 \times 10^{-5}\). Our training dataset is sourced from MOSEI\cite{mosei} and AVSPEECH\cite{avspeech}, which contain half-body speech scenarios. Additionally, we enhance our dataset by collecting videos from the internet, resulting in a total duration of approximately 275 hours.

\subsection{Hand Motion Generation Comparisons}
We compared our methods to other baseline motion generation methods using the Talkshow dataset, focusing primarily on hand movements. Since other methods mostly operate on SMPL poses, we first extracted hand translations from the generated SMPL poses before calculating hand movement metrics for these methods. To demonstrate that directly generating hand motions with our methods outperforms all SMPL poses-based methods in terms of hand movement, we also trained our method using the same setup with SMPL poses. Each method was tested on 100 input audios from the Talkshow test dataset, and for each input audio, 50 results were generated. For metrics, DIV is used to measure the diversity of the generated results, BA describes the beat alignment between audio and the results, PCK is the percentage of results close to the ground truth (GT) motion, and Fréchet Gesture Distance (FGD) represents the distance between the distribution of generated motion and GT motion. 

\autoref{tab:quantitative_comp_motion} demonstrates that our MANO version method leads by a large margin regarding DIV, which we believe relates most to the vividness and expressiveness of hand motion, whereas other SMPL-based methods often generate monotonous results. Despite various starting gestures, these methods tend to move hands in front of the chest or stay in the first place. In terms of BA, our MANO-based method also shows superiority. These two metrics play the most important role in providing a vivid and expressive co-speech driving signal for the next stage of video generation. Our method's lower scores in PCK and FGD are expected, as the hand movements from other methods are calculated based on the forward winematics of SMPL, thus restricted to be distributed closer to the GT hand movements. Our MANO-based method, in contrast, has more freedom to move hands differently from the GTs. Qualitative results also show that discarding SMPL forward kinematics produces more flexibility without compromising the reasonability of the generation. We also compared our SMPL-based method with baselines in SMPL poses, and our method still holds a distinct lead in diversity. In \autoref{fig:hand distribution}, the 2D hand position distribution generated by various hand motion generation methods is illustrated. Our method is capable of producing positions with a wider range and movements in more diverse patterns.

\FloatBarrier

\begin{figure*}[tb]
  \centering
  \includegraphics[width=0.7\textwidth]{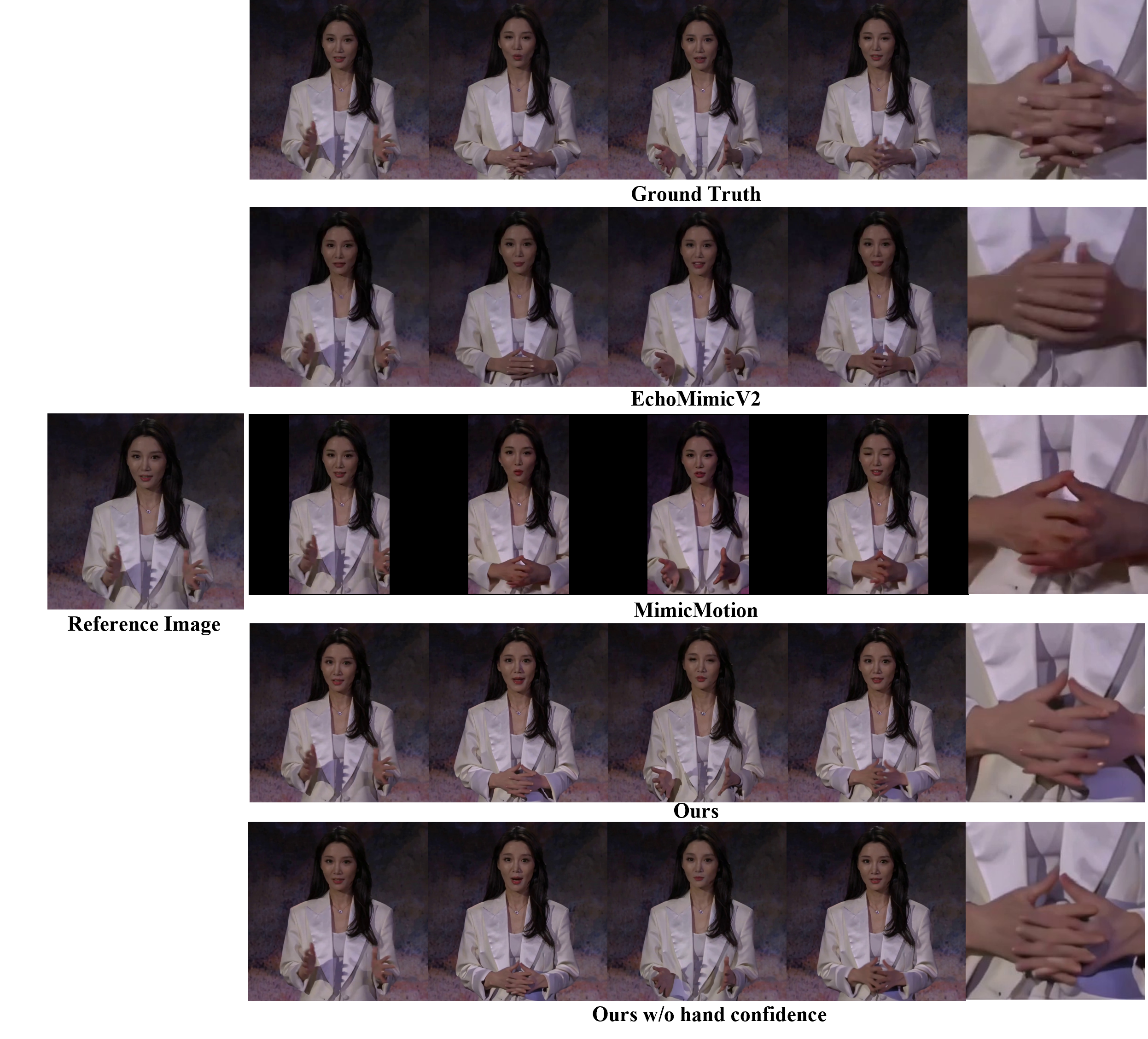}
  \caption{The qualitative comparisons with pose-driven body animation methods, based on the EMTD dataset.
  }
  \label{fig:quality_emtd}
\end{figure*}

\subsection{Video Generation Comparisons}

We conduct the upper body animation comparisons on the EMTD\cite{echomimicv2} dataset. To demonstrate the superiority of our proposed method, we evaluate the models using several metrics. We employ Fréchet Inception Distance (FID)\cite{fid}, SSIM\cite{ssim}, and PSNR\cite{psnr} to assess the quality of the generated frames. Fréchet Video Distance (FVD)\cite{fvd} is used to gauge the overall coherence of the generated videos. To evaluate identity consistency, we calculate the cosine similarity (CSIM) between the facial features of the reference image and the generated video frames. We also utilize Sync-C, as proposed by SyncNet\cite{syncnet}, to assess the synchronization quality between lip movements and audio signals. Furthermore, we measure Hand Keypoint Confidence (HKC) to evaluate the quality of hand representation in generated frames, while Hand Keypoint Variance (HKV) serves as an indicator of the richness of hand movement. Additionally, EFID\cite{tian2025emo} is adopted to quantitatively assess the divergence in expressions between the synthesized videos and those in the ground truth dataset.

There are a few works that have achieved audio-driven body animation; however, some, such as DiffTED\cite{diffted}, have not open-sourced their pre-trained models, while others like Vlogger\cite{vlogger} and CyberHost\cite{cyberhost} have only released demo videos. To facilitate comparisons with other state-of-the-art methods, we compare our approach with EchoMimicV2\cite{echomimicv2}, which animates the hands using pre-extracted hand pose sequences while driving the rest of the body through audio. We also compare our work with MimicMotion\cite{mimicmotion2024}, a video-driven body reenactment method. Both EchoMimicV2 and MimicMotion rely on the pose sequences from the ground truth to animate the body. Since CyberHost and Vlogger have not open-sourced their models, we conduct relevant comparison experiments based on demo videos retrieved from their respective homepages.

As illustrated in \autoref{tab:quantitative_comp}, "w/o motion gen" denotes that our model utilizes the pose sequence from the ground truth instead of the motion generation results, compared to EchoMimicV2 and MimicMoton, the results demonstrate an advantage in video quality assessment, as evidenced by the lower FVD scores. Additionally, our method outperforms others in terms of individual frame quality, as indicated by improved image quality (FID, SSIM, PSNR) scores. Compared "w/o motion gen" with other validation sets, using the original pose as a driver will naturally make the generated results consistent with GT, thereby improving the quality metrics of images and videos, however, higher HKV denotes that our model could generate more diverse motion sequence. Meanwhile, our models have the ability to maintain the identity proved by the CSIM. Lower EFID proved that our model could generate more lively facial expressions. Besides, as illustrated in \autoref{fig:quality_emtd}, our model could generate hands with clear structure and logical interaction.

\begin{figure*}[tb]
  \centering
  \includegraphics[width=0.7\textwidth]{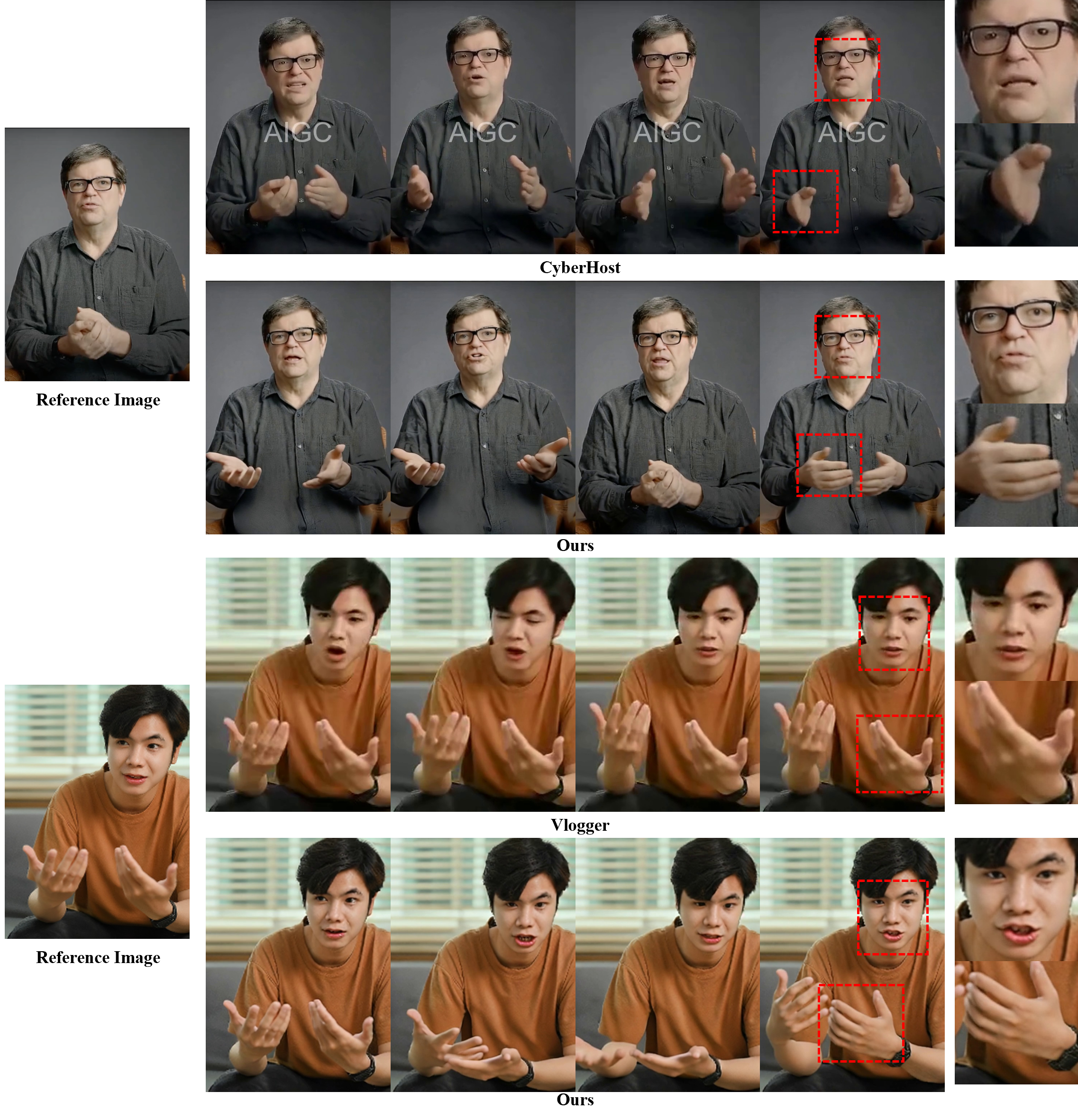}
  \caption{The qualitative comparisons with audio-driven body animation methods.
  }
  \label{fig:quality_demo}
\end{figure*}

We also compared our model with CyberHost and Vlogger based on demos from their homepages. The results in \autoref{tab:quantitative_comp} show that, since most of Vlogger's demos exhibit minimal motion (indicated by very low HKV), the HKC is relatively high. However, proved by the highest HKV, our model has the ability to generate much more diverse motion than other methods. The quality comparisons are shown in \autoref{fig:quality_demo}, our method could generate frames with higher quality, while better preserve the original facial features.

\section{Conclusion}
This paper devises a two-stage framework for co-speech human video generation based on diffusion models. In this framework, EMO \cite{tian2025emo} is extended to enable the generation of holistic facial expressions and upper body motions. We introduce the concept of  \textbf{"pixels prior IK"} , because we find that hand movements are the most correlated with audios among the different body parts, and the video generation backbone naturally incorporates human body Inverse Kinematics priors in form of pixels. Consequently, in Stage 1, only hand movements are generated and subsequently used as control signals in Stage 2. The results demonstrate that this framework can produce more expressive and vivid human videos compared to other methods. We hope that this work can offer a new thought for audio-driven talking head animation.

\FloatBarrier
\bibliography{main}
\bibliographystyle{icml2025}




\end{document}